\documentclass[sigconf]{acmart}
\usepackage{multirow}
\usepackage{pifont}
\AtBeginDocument{%
  }

\copyrightyear{2026}
\acmYear{2026}
\setcopyright{cc}
\setcctype{by}
\acmConference[MM '26]{Proceedings of the 34th ACM International Conference on Multimedia}{November 10--14, 2026}{Rio de Janeiro, Brazil}
\acmBooktitle{Proceedings of the 34th ACM International Conference on Multimedia (MM '26), November 10--14, 2026, Rio de Janeiro, Brazil}
\acmDOI{10.1145/3767308.3835514}
\acmISBN{979-8-4007-2213-4/2026/11}

\begin{document}

\title{GenPrior: Unleashing Text-to-Motion Generative Priors for Zero-Shot Skeleton-based Action Recognition}
\author{Jidong Kuang}
\orcid{0009-0000-4755-6055}
\affiliation{%
  \institution{School of Cyber Science and Engineering, Southeast University}
  \city{Nanjing}
  \country{China}}
\email{jidongkuang@seu.edu.cn}

\author{Hongsong Wang}
\correspondingauthor
\orcid{0000-0002-9464-1778}
\affiliation{%
  \institution{School of Computer Science and Engineering, Southeast University}
  \institution{Key Laboratory of New Generation Artificial Intelligence Technology and Its Interdisciplinary Applications}
  \city{Nanjing}
  \country{China}}
\email{hongsongwang@seu.edu.cn}

\author{Jie Gui}
\correspondingauthor
\orcid{0000-0002-9450-1759}
\affiliation{%
  \institution{School of Cyber Science and Engineering, Southeast University}
  \institution{Purple Mountain Laboratories}
  \institution{Engineering Research Center of Blockchain Application, Supervision and Management}
  \city{Nanjing}
  \country{China}}
\email{guijie@seu.edu.cn}

\renewcommand{\shortauthors}{Kuang et al.}
\begin{abstract}

  Zero-shot skeleton-based action recognition (ZSAR) aims to recognize unseen action categories by aligning skeleton features with textual semantics. However, existing methods rely on text-derived prototypes that inherently lack geometric structure and physical constraints, resulting in a pronounced \textit{semantic-kinematic gap}. To bridge this gap, we propose \textbf{GenPrior}, the first framework to exploit generative priors from pre-trained Text-to-Motion (T2M) models for ZSAR. Specifically, we introduce Dispersion-Gated Feature Fusion, which distills kinematic prototypes and intra-class dispersion from generative motion sequences and employs a learned gating network to adaptively inject reliable structural cues into textual embeddings while suppressing synthetic artifacts. Furthermore, we propose Generative Prototype Refinement, which leverages these generation-enhanced prototypes as anchors to mine high-confidence unseen samples, calibrating class prototypes toward the true distribution and thereby unleashing strong performance gains. Extensive experiments on NTU-60, NTU-120, and PKU-MMD demonstrate that GenPrior achieves state-of-the-art performance under both zero-shot and generalized zero-shot settings. Code is available at \url{https://github.com/jidongkuang/GenPrior}.
\end{abstract}



\maketitle

\section{Introduction}

Human action recognition has long been a central topic in computer vision and plays an important role in a wide range of applications, including health monitoring~\cite{xie2025harmony}, medical rehabilitation~\cite{yu2022egcn}, human-computer interaction~\cite{narayana2018gesture}, and robotics~\cite{wang2025recognizing}. Compared with RGB videos, skeleton sequences explicitly preserve the spatiotemporal topology of human joints and motion trajectories, offering superior privacy protection, more compact representations, and greater robustness to background clutter. As a result, they have gradually become an important modality for action understanding~\cite{wang2025foundation, weng2025usdrl}. Nevertheless, most existing skeleton-based action recognition methods~\cite{wang2025heterogeneous} rely heavily on sufficient class-wise annotated data, which is often unrealistic in open-world scenarios. Acquiring large-scale labeled training samples is typically expensive, if not infeasible. To address this, Zero-Shot Skeleton-based Action Recognition (ZSAR) has emerged, aiming to recognise unseen action classes via class semantics while relying solely on seen-class samples during training~\cite{zhou2023zero, li2024sa, chen2024fine, zhu2024part, zhu2025semantic, zhu2026boosting}.

\begin{figure}[t]
    \centering
    \includegraphics[width=\linewidth]{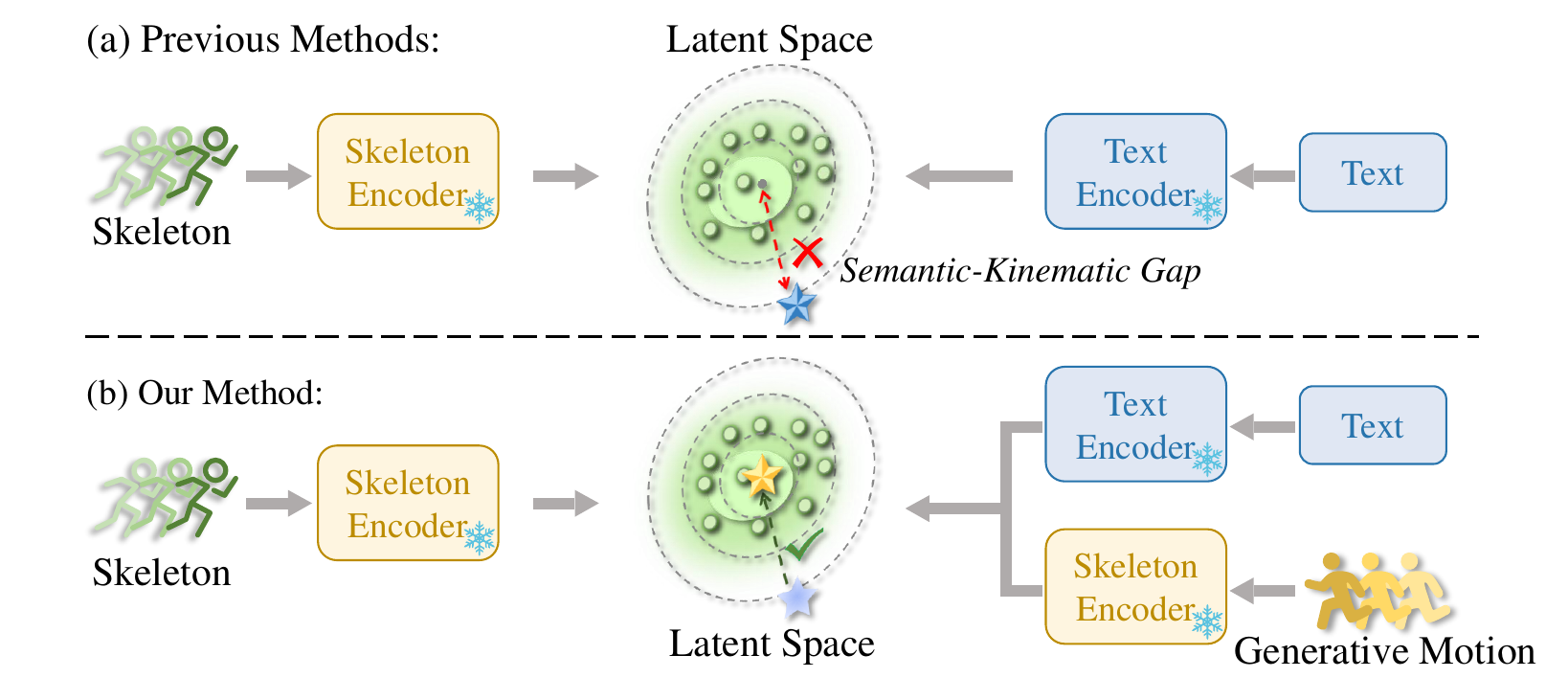} 
    \caption{Overview of our method versus previous ZSAR methods. \textbf{(a) Previous methods} rely on text-derived prototypes, which lack physical constraints and are biased toward seen domains, resulting in a pronounced \textit{semantic-kinematic gap}. \textbf{(b) Our method} introduces generative motions from Text-to-Motion generation models as motion priors to enrich textual semantics and yield prototypes that better align with the unseen distribution.}
    \label{fig:intro_framework}
    \vspace{-0.4cm}
\end{figure}

Existing ZSAR methods are generally built upon a cross-modal alignment paradigm, where skeleton features and class textual descriptions are projected into a shared semantic space, and unseen-class recognition is performed through distance matching or prototype retrieval. Although such methods alleviate the need for large-scale labeled data to some extent, their performance remains strongly constrained by the quality of semantic representations. The fundamental reason is that textual descriptions primarily emphasize semantic concepts, whereas skeleton sequences encode the geometric structure and dynamic patterns of human motion, leading to a pronounced semantic-kinematic gap. In particular, for unseen classes, semantic prototypes constructed solely from class names or static text embeddings usually lack the structural constraints required by real motions, making it difficult to faithfully characterize their actual distribution in the skeleton feature space.
Meanwhile, the learned prototypes and decision boundaries in existing methods are largely shaped by supervision from seen classes, while the text-derived prototypes themselves lack structural kinematic support for the real motion patterns of unseen classes. As a result, they are prone to systematic shifts when confronted with unseen samples at test time.

Meanwhile, recent advances in Text-to-Motion (T2M) generation have opened up new opportunities for addressing these challenges. Modern T2M models are able to generate semantically consistent, temporally coherent, and kinematically plausible skeleton motion sequences from natural language descriptions, demonstrating strong capability in text-motion alignment, motion realism, and diversity~\cite{guo2022generating, zhang2024motiondiffuse, zhang2023remodiffuse, lu2025scamo}. These developments suggest that T2M models are not merely motion synthesizers, but also implicit providers of structural motion priors for action categories, thereby offering a new avenue for compensating for the lack of motion-aware constraints in textual semantics for ZSAR.

To further improve the generalization ability of ZSAR, two objectives must be addressed simultaneously: enriching semantic prototypes with missing structural motion priors, and refining decision boundaries with more reliable unseen class prototypes. Motivated by this observation, and as illustrated in Figure~\ref{fig:intro_framework}, we propose \textbf{GenPrior}, whose key idea is to unleash the generative priors embedded in pre-trained Text-to-Motion models and exploit them for semantic enhancement and prototype modeling of unseen classes. By introducing kinematically grounded generative information, GenPrior provides a new form of structural support for conventional text-derived zero-shot skeleton-based action recognition.

Specifically, we first incorporate the class-level structural information contained in generated motions into semantic representations in an adaptive manner. For each class, we generate diverse motions and extract their features using a pre-trained skeleton encoder, from which we estimate class-level generative kinematic prototypes and intra-class dispersion. We then introduce Dispersion-Gated Feature Fusion to dynamically regulate the integration of generative priors. This design allows reliable structural motion cues to be selectively incorporated while preserving the dominance of textual semantics, thereby yielding robust generation-enhanced semantic features. These enhanced representations are subsequently aligned with real skeleton features through variational encoding in a shared latent space. Furthermore, to alleviate the decision bias induced by seen-class supervision, we introduce Generative Prototype Refinement at inference time. Using the generative prototypes as initialization anchors, we mine high-confidence support samples and refine the class prototypes accordingly, driving them toward the true distribution of unseen classes.

By using generated motions to structurally guide the semantic space and real motions to refine generation-based prototypes, GenPrior effectively mitigates domain-shift artifacts and substantially improves the adaptability of decision boundaries to unseen classes. Extensive experiments on NTU-60, NTU-120, and PKU-MMD demonstrate that GenPrior achieves state-of-the-art performance under both zero-shot and generalized zero-shot settings.

Our contributions are summarized as follows:
\begin{itemize}
\item We propose \textbf{GenPrior}, the first framework to exploit generative priors from T2M models for ZSAR, thereby compensating for the lack of structural motion information in conventional text-derived prototypes.
\item We introduce a mechanism that uses generative priors to enrich semantic representations and refine class prototypes, effectively alleviating the semantic-kinematic gap.
\item Extensive experiments on NTU-60, NTU-120, and PKU-MMD show that GenPrior achieves state-of-the-art performance under both ZSL and GZSL settings.
\end{itemize}

\begin{figure*}[t]
    \centering
    \includegraphics[width=0.85\textwidth]{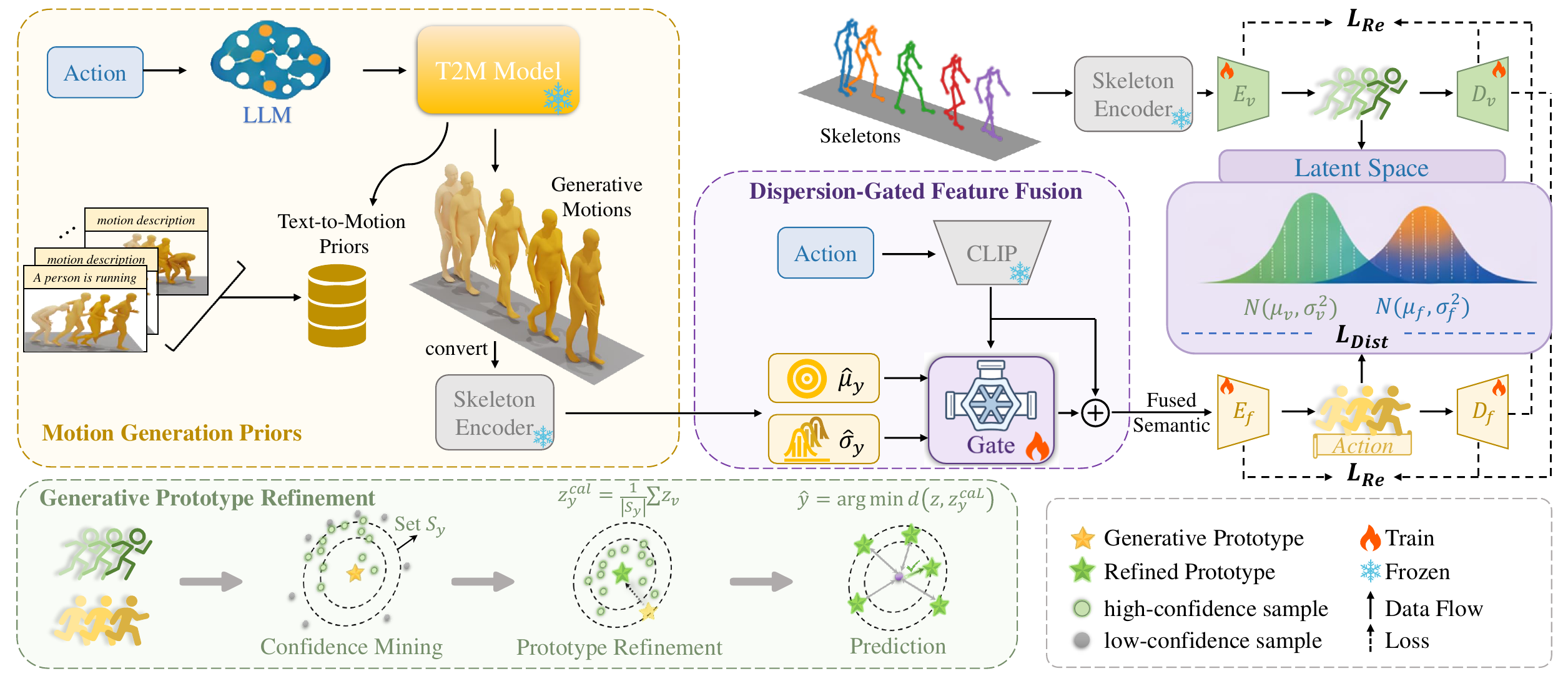}
    \caption{
    \textbf{Overall architecture of GenPrior.}
    A frozen T2M model generates motions to derive motion priors.
These priors are adaptively fused with textual representations to form generation-enhanced semantics, which are aligned with real skeleton features in a shared latent space via VAE encoding.
At inference, generative prototype refinement mines high-confidence samples to iteratively update class prototypes, enabling adaptive calibration for unseen categories.
    }
    \label{fig:framework}
\end{figure*}

\section{Related Work}
\noindent \textbf{Zero-shot Skeleton-based Action Recognition:}
ZSAR aims to recognize human actions from skeleton sequences without access to labeled visual samples of unseen categories during training. The core challenge is to bridge the \textit{semantic-kinematic gap} between low-level 3D joint trajectories and high-level textual descriptions. Existing methods can be broadly grouped into two paradigms: discriminative embedding alignment and latent generative alignment.
Discriminative methods project skeleton features and textual semantics into a shared latent space for nearest-neighbor retrieval. Early works such as ReViSE~\cite{hubert2017learning} and JPoSE~\cite{wray2019fine} establish cross-modal alignment through metric learning. Subsequent methods improve temporal modeling and semantic expressiveness. For example, SMIE~\cite{zhou2023zero} maximizes mutual information across modalities, and DVTA~\cite{kuang2025zero} introduces dual alignment with cross-attention. More recent works leverage large language models to enrich class semantics, including part-aware descriptions in PURLS~\cite{zhu2024part} and STAR~\cite{chen2024fine}, multi-sentence contrastive alignment in InfoCPL~\cite{xu2025information}, and multi-turn semantic expansion in Neuron~\cite{chen2025neuron}. Related directions also include prompt learning in SCoPLe~\cite{zhu2025semantic}, prototype-guided alignment in PGFA~\cite{zhou2025zero}, and training-free test-time adaptation in Skeleton-Cache~\cite{zhu2026boosting}. Despite these advances, such methods remain largely grounded in linguistic representations, which lack explicit geometric structure and physical constraints, limiting their robustness for complex unseen motions.
A second line of work addresses seen-class bias through latent generative alignment, typically by encoding visual and semantic modalities into a shared Gaussian space. SynSE~\cite{gupta2021syntactically} introduces verb-noun compositional embeddings, SA-DVAE~\cite{li2024sa} disentangles class semantics from motion style variation, and FS-VAE~\cite{wu2025frequency} incorporates frequency-domain analysis to capture multi-scale temporal patterns. More recently, stronger generative architectures have been explored beyond VAEs. TDSM~\cite{do2025bridging} aligns features through text-conditioned denoising diffusion, while Flora~\cite{chen2025learning} models open-form flows with neighbor-aware semantics to construct more robust decision boundaries.

Unlike previous works, our method introduces a different paradigm. Instead of training constrained internal generators or relying solely on linguistic representations, we exploit external pre-trained Text-to-Motion models as motion prior sources.

\noindent \textbf{Text-to-Motion Generation: }
Text-to-motion (T2M) generation aims to synthesize realistic 3D human motion sequences from natural language descriptions and has advanced rapidly in recent years. Early work such as T2M~\cite{guo2022generating} introduced a temporal VAE framework and established HumanML3D as a standard benchmark for this task. Since then, the field has evolved toward more powerful probabilistic generative models.
One major line of research explores vector-quantized discrete modeling~\cite{guo2022tm2t,yuan2024mogents,pinyoanuntapong2024bamm,lu2025scamo}. Methods such as T2M-GPT~\cite{zhang2023generating} and MoMask~\cite{guo2024momask} compress continuous motion into discrete codebooks and formulate motion generation as token prediction. In parallel, diffusion-based methods have become the dominant paradigm~\cite{kim2023flame,yuan2023physdiff,zhou2024emdm,zhang2025energymogen}. Representative works such as MotionDiffuse~\cite{zhang2024motiondiffuse} and MDM~\cite{tevet2022human} show that text-conditioned denoising can generate high-fidelity motion sequences, while later models including MLD~\cite{chen2023executing} and StableMoFusion~\cite{huang2024stablemofusion} further improve efficiency by performing diffusion in compact latent spaces.

Most of these models are trained on HumanML3D~\cite{guo2022generating}, a large-scale motion--text benchmark built upon motion capture data from sources such as AMASS~\cite{mahmood2019amass}. The broad action coverage of this dataset enables T2M models to learn a rich mapping from textual semantics to plausible motion patterns. Building on this capability, we employ a pre-trained T2M model to provide kinematically grounded guidance for action categories.

\noindent \textbf{Generative Zero-Shot Learning: }
Generative zero-shot learning is an important branch of zero-shot learning that addresses unseen-class recognition by synthesizing visual features from semantic descriptions, thereby converting zero-shot recognition into a supervised classification problem. Early work such as f-CLSWGAN~\cite{xian2018feature} introduced conditional Wasserstein GANs to generate discriminative visual features. CADA-VAE~\cite{schonfeld2019generalized} instead aligned visual and semantic modalities in a shared latent space through cross-modal variational learning, while f-VAEGAN-D2~\cite{xian2019f}, IZF~\cite{shen2020invertible}, and TF-VAEGAN~\cite{narayan2020latent} further improved feature synthesis by combining VAEs, GANs, normalizing flows, and semantic feedback. Subsequent methods, including FREE~\cite{chen2021free}, HSVA~\cite{chen2021hsva}, CE-GZSL~\cite{han2021contrastive}, and DSP~\cite{chen2023evolving}, enhanced generative ZSL from the perspectives of feature refinement, hierarchical adaptation, contrastive supervision, and dynamic prototype evolution. More recently, diffusion-based generative ZSL has also emerged, as exemplified by DIG-ZSL~\cite{fu2024discriminative} and ZeroDiff~\cite{ye2025zerodiff}.

Although generative zero-shot learning has achieved strong results in 2D image recognition, its paradigm of treating synthesized features as training data does not transfer well to 3D skeleton-based recognition, where structural plausibility and motion continuity are crucial. This limitation motivates us to use generated motions as priors rather than raw training samples.

\section{Preliminaries}

\noindent \textbf{Problem Setup: }
Let $\mathcal{D} = \{(X_i, y_i)\}_{i=1}^N$ denote a skeleton action dataset, where each skeleton sequence $X_i \in \mathbb{R}^{C \times T \times V \times M}$ consists of $C$-dimensional joint coordinates across $T$ frames, $V$ joints, and $M$ body subjects, with $y_i \in \mathcal{Y}$ being the corresponding action label. Each class $y \in \mathcal{Y}$ is associated with a semantic description $l_y$, forming the semantic set $\mathcal{L} = \{l_y\}_{y \in \mathcal{Y}}$. The dataset is split into a seen-class training set $\mathcal{D}^s_{\mathrm{tr}}$, a seen-class test set $\mathcal{D}^s_{\mathrm{te}}$, and an unseen-class test set $\mathcal{D}^u_{\mathrm{te}}$, where $\mathcal{Y} = \mathcal{Y}^s \cup \mathcal{Y}^u$ and $\mathcal{Y}^s \cap \mathcal{Y}^u = \emptyset$. Only $\mathcal{D}^s_{\mathrm{tr}}$ is available during training. At inference, the model is evaluated on $\mathcal{D}^u_{\mathrm{te}}$ in the ZSL setting and on $\mathcal{D}^s_{\mathrm{te}} \cup \mathcal{D}^u_{\mathrm{te}}$ in the GZSL setting. We further introduce a pre-trained Text-to-Motion (T2M) model $\mathcal{G}$ to provide generative priors that reduce the semantic-kinematic gap.

\noindent \textbf{Text-to-Motion Generation: }
Text-to-Motion (T2M) generation synthesizes a motion sequence $\hat{X}\in\mathbb{R}^{T' \times V \times D}$ from a textual description $c$ (e.g., ``a person waves their right hand''), where $T'$ denotes the number of generated frames, $V$ the number of joints, and $D$ the relative joint feature dimension encoding local velocities and relative rotations. Modern T2M models typically adopt conditional diffusion models that denoise motion features conditioned on text embeddings. The generation process is written as:
\begin{equation}
    \hat{X} = \mathcal{G}(c, \epsilon), \quad \epsilon \sim \mathcal{N}(0, I),
\end{equation}
where the stochasticity introduced by $\epsilon$ enables sampling multiple motion sequences that are semantically consistent yet kinematically diverse from the same textual condition $c$.

T2M models are typically pre-trained on large-scale motion-text datasets such as HumanML3D~\cite{guo2022generating}, endowing them with two properties that are central to our method. First, they generalize across categories and can generate plausible motions from open-vocabulary textual descriptions, including novel action compositions unseen during training. Second, they preserve human skeletal topology and motion dynamics, thereby providing geometric structure that is absent from pure textual embeddings.

\section{Method}
An overview of the proposed framework is shown in Fig.~\ref{fig:framework}.

\subsection{Generation Enhanced Semantic Features}

\noindent \textbf{Motion Generation Priors: }

In our framework, we formulate the pre-trained T2M model $\mathcal{G}$ as a frozen external motion prior source. For each action class $y \in \mathcal{Y}$, we first generate a diverse set of textual descriptions $\{c_y^{(j)}\}_{j=1}^J$ via a large language model, and then feed each description into $\mathcal{G}$ with $K$ random seeds per description. This process yields a total of $N_g = J \times K$ generated motion sequences $\{\hat{X}_y^{(n)}\}_{n=1}^{N_g}$ per class. 
To bridge the representation format, these generated motions are converted from relative spatial configurations to absolute joint coordinates. Each converted synthetic sequence is subsequently passed through a pre-trained skeleton encoder $\Phi(\cdot)$ to extract its feature representation:
\begin{equation}
    \hat{f}_y^{(n)} = \Phi(\hat{X}_y^{(n)}).
\end{equation}

From this synthetic feature set, we distill two critical statistics. The first is the \textit{generative kinematic prototype}, defined as the centroid of the synthetic features for each class:
\begin{equation}
    \hat{\mu}_y = \frac{1}{N_g} \sum_{n=1}^{N_g} \hat{f}_y^{(n)},
\end{equation}
which encodes the central kinematic structure of category $y$ in the motion feature space. The second is the intra-class dispersion vector, which captures the per-dimension standard deviation of synthetic features within each class:
\begin{equation}
    \hat{\sigma}_y = \sqrt{\frac{1}{N_g} \sum_{n=1}^{N_g} \left(\hat{f}_y^{(n)} - \hat{\mu}_y\right)^2},
\end{equation}
where the square root and squaring are element-wise. The dispersion vector $\hat{\sigma}_y$ reflects the variability of features generated by the T2M model for category $y$ at each feature dimension.

\noindent \textbf{Dispersion-Gated Feature Fusion: }
Although generative priors encode rich kinematic structure, they inevitably contain synthetic artifacts such as joint jitter, physical implausibility, and motion incoherence. Directly using them as real features for classifier training would introduce domain noise rather than mitigate the bias between seen and unseen classes. We therefore propose a learnable dispersion-gated mechanism that converts generative priors into textual embeddings.

Concretely, for each class $y$, we extract its textual feature $e_y = \Psi(l_y)$ using a pre-trained text encoder $\Psi(\cdot)$, and project the generative kinematic prototype $\hat{\mu}_y$ into the textual embedding space via a learnable linear layer, yielding $p_y$. We design a gating network to control the incorporation of generative information by jointly considering the textual semantics, the projected prototype, and the dispersion pattern of the generated features. Specifically, the dispersion vector $\hat{\sigma}_y$ is first mapped into the textual embedding space via a linear projection $W_\sigma$, and then concatenated with $e_y$ and $p_y$ to produce a scalar gate:
\begin{equation}
    g_y = \sigma\!\left(f_\theta\!\left([e_y \parallel p_y \parallel W_\sigma \hat{\sigma}_y] \right)\right),
\end{equation}
where $\parallel$ denotes concatenation, $f_\theta$ is a learnable feedforward network, and $\sigma(\cdot)$ denotes the sigmoid function. The gate value is bounded in $(0,1)$, where a larger value allows more generative information to be incorporated into the textual representation, while a smaller value suppresses incorporation when the generated features are less reliable or less informative. By conditioning on all three inputs, the gate captures both the quality of the generated features reflected by $\hat{\sigma}_y$ and the semantic and kinematic complementarity between the textual and generative modalities. The fused semantic representation is then computed as:
\begin{equation}
    \tilde{e}_y = e_y + g_y \cdot p_y,
\end{equation}
where the gate $g_y$ uniformly scales the projected prototype $p_y$. With the fused semantic representations $\tilde{e}_y$, we encode both 
modalities into a shared latent space for cross-modal alignment. 
For a skeleton sequence $X_i$, we extract its feature 
$f_v = \Phi(X_i)$ and map both $f_v$ and $\tilde{e}_y$ into 
latent distributions via their respective variational encoders:
\begin{equation}
    z_v \sim \mathcal{E}_v(z \mid f_v) = \mathcal{N}(\mu_v, \sigma_v^2), 
    \quad 
    z_f \sim \mathcal{E}_f(z \mid \tilde{e}_y) = \mathcal{N}(\mu_f, \sigma_f^2),
\end{equation}
where $z_v$ and $z_f$ denote the visual and semantic latent variables, 
and $\mu_v, \sigma_v^2$, $\mu_f, \sigma_f^2$ are the corresponding 
mean and variance vectors.

\subsection{Generative Prototype Refinement}
Classifiers trained solely on seen-class data inherently exhibit a decision bias toward seen categories. Conventional calibration strategies typically rely on preset thresholds or entropy-based gating to balance the prediction distribution, yet they lack direct access to the distribution of unseen classes.

To address this issue, we leverage generative priors as anchors that provide direct visual references for calibrating unseen-class decision boundaries. The key observation is that semantic prototypes derived solely from textual encodings may fail to faithfully capture the true unseen-class distribution in the latent space. By mining high-confidence test samples under the guidance of these generative anchors, we progressively refine the prototypes toward the real data distribution.

\noindent\textbf{Confidence-Based Sample Mining:}
All test samples are first encoded into the latent space and matched to class prototypes via nearest-neighbor retrieval:
\begin{equation}
    \hat{y}_i = \arg\min_{y} \, d(z_v^{(i)},\; z_f^y),
\end{equation}
where $d(\cdot,\cdot)$ denotes the Euclidean distance and $z_f^y$ is the semantic latent prototype of class $y$. This initial assignment associates each test sample with its most likely class. For each assigned group, we estimate sample confidence by converting the negative distance vector into a softmax distribution and computing its information concentration:
\begin{equation}
    H_i = -\sum_{y} p^{(i)} \log p^{(i)}, \quad \text{where} \quad p^{(i)} = \frac{\exp(-d(z_v^{(i)}, z_f^y))}{\sum_{y'}\exp(-d(z_v^{(i)}, z_f^{y'}))}.
\end{equation}
A lower $H_i$ indicates higher confidence, as the sample's distance distribution is more peaked toward a single class. Within each class $y$, we rank all assigned samples by ascending $H_i$ and select the top $\eta$-fraction as the high-confidence support set $\mathcal{S}_y$, where $\eta \in (0, 1]$ controls the selection ratio.

\noindent\textbf{Prototype Refinement:}
Using the mined support set, we compute the refined prototype of each class as the mean of its selected high-confidence samples:
\begin{equation}
    z_y^{\mathrm{cal}} = \frac{1}{|\mathcal{S}_y|}\sum_{z_v^{(i)} \in \mathcal{S}_y} z_v^{(i)}.
\end{equation}
If no high-confidence sample is mined for a particular class $y$, that is, when $|\mathcal{S}_y| = 0$, the prototype falls back to the semantic latent representation $z_f^y$. Although the refined prototypes are constructed entirely from test features, their reliability is supported by the initial assignment guided by generative priors.

\subsection{Training and Inference}

\noindent\textbf{Training: }

The alignment objective consists of two terms. The first term is intra modal and cross modal reconstruction:
\begin{equation}
    \mathcal{L}_{\mathrm{Re}} = \sum_{k \in \{v,f\}} \mathbb{E}_{\mathcal{E}_k(z_k \mid x_k)} \left[ \log \mathcal{D}_k(x_k \mid z_k) + \log \mathcal{D}_{\bar{k}}(x_{\bar{k}} \mid z_k) \right],
\end{equation}
where $x_v = f_v$, $x_f = \tilde{e}_y$, $\bar{k}$ denotes the opposite modality of $k$, and $\mathcal{D}$ is modality specific decoder. The second term is distribution alignment, which explicitly constrains the geometric consistency of the two latent distributions:
\begin{equation}
    \mathcal{L}_{\mathrm{Dist}} = \| \mu_v - \mu_f \|_2^2 + \| \sigma_v^2 - \sigma_f^2 \|_2^2.
\end{equation}
This objective encourages tight alignment between the visual and semantic distributions in the latent space, thereby reducing the modality gap while preserving modality specific information. The overall cross modal alignment loss, with $\lambda_{\mathrm{Dist}}$ controlling the contribution of the distribution alignment term, is defined as
\begin{equation}
    \mathcal{L}_{\mathrm{Align}} = \mathcal{L}_{\mathrm{Re}} + \lambda_{\mathrm{Dist}} \mathcal{L}_{\mathrm{Dist}}.
\end{equation}

\noindent\textbf{Inference: }
For each test skeleton sequence $X_{\mathrm{te}}$, we extract its skeleton feature $f_v = \Phi(X_{\mathrm{te}})$ and obtain its latent representation $z_v$ through the variational encoder $\mathcal{E}_v$. The Generative Prototype Refinement module calibrates the unseen class prototypes and yields $z_y^{\mathrm{cal}}$. The prediction is given by
\begin{equation}
    \hat{y} = \arg\min_{y \in \mathcal{Y}^u} d(z_v, z_y^{\mathrm{cal}}).
\end{equation}

In the generalized setting, the search space expands to $\mathcal{Y}^s \cup \mathcal{Y}^u$. After obtaining the calibrated prototypes $z_y^{\mathrm{cal}}$, we compute for each test sample the minimum distance to the seen and unseen domains, respectively:
\begin{equation}
    \delta_s = \min_{y \in \mathcal{Y}^s} d(z_v, z_y^{\mathrm{cal}}), \quad
    \delta_u = \min_{y \in \mathcal{Y}^u} d(z_v, z_y^{\mathrm{cal}}).
\end{equation}
Their ratio indicates whether the sample is more likely to belong to the seen or unseen domain. The unified prediction is formulated as
\begin{equation}
    \hat{y} = \arg\min_{y \in \mathcal{Y}^s \cup \mathcal{Y}^u} \left[ d(z_v, z_y^{\mathrm{cal}}) + \lambda \cdot \mathbb{I}\left[(y \in \mathcal{Y}^s) \oplus (\delta_s / \delta_u \leq \gamma)\right] \right].
\end{equation}
Here, $\gamma$ is a calibration threshold, $\oplus$ denotes the exclusive or operator, and $\lambda$ is a sufficiently large constant used to mask candidates from the opposite domain. When a sample is classified as seen ($\delta_s / \delta_u \leq \gamma$), the penalty $\lambda$ is applied to all unseen candidates, thereby restricting the search to seen classes.

\section{Experiments}
\subsection{Datasets}
\noindent \textbf{NTU RGB+D 60}~\cite{shahroudy2016ntu}: Captured by three Microsoft Kinect v2 cameras from different angles, this is a widely adopted dataset for skeleton-based human action recognition. It comprises 60 action categories performed by one or two subjects, totaling 56,880 skeleton sequence samples. Each frame provides the 3D coordinates of 25 body joints.

\noindent \textbf{NTU RGB+D 120}~\cite{liu2019ntu}: As an extension of NTU RGB+D 60, this dataset expands the action categories to 120 and adds 57,367 new skeleton sequences, yielding a total of 114,480 samples.

\noindent \textbf{PKU-MMD}~\cite{liu2017pku}: This dataset encompasses 51 action categories with nearly 20,000 skeleton sequences. It shares the same 25-joint skeleton topology as the NTU datasets.

\subsection{Implementation Details}
For data preprocessing, we follow the protocol established in Cross-CLR~\cite{li20213d}. To construct motion priors, we use GPT-4 to expand each action class name into $J=40$ human-centered motion descriptions, which are then fed into the StableMoFusion~\cite{huang2024stablemofusion} T2M model. Sampling $K=10$ random seeds per description yields $N_g=400$ synthetic motion sequences for each class. We use Shift-GCN~\cite{cheng2020skeleton} to extract both real and generated skeleton features, and adopt a pre-trained CLIP model (ViT-L/14@336px) to encode textual features. Our framework is implemented in PyTorch and optimized with AdamW, with an initial learning rate of $1 \times 10^{-4}$ and a weight decay of $1 \times 10^{-2}$. The batch size is set to 128. All experiments are conducted on a single NVIDIA GeForce RTX 4090 GPU.

Regarding the experimental splits, we adopt the standard splits from SynSE~\cite{gupta2021syntactically} for Zero-Shot Learning (ZSL) and conduct the Generalized Zero-Shot Learning (GZSL) evaluation on the same splits. Under the GZSL setting, we report the Seen accuracy (S), Unseen accuracy (U), and their Harmonic mean (H = (2 × S × U) / (S + U)). Furthermore, following the random split setting from SA-DVAE~\cite{li2024sa}, we perform three random category splits for each dataset and report the averaged results to mitigate biases introduced by specific split configurations.

\begin{table}[t]
\centering
\caption{Zero-shot action recognition accuracy (\%) on NTU-60 and NTU-120. The best results are highlighted in bold, while the second-best results are underlined. GenPrior and SC~\cite{zhu2026boosting} leverage unlabeled test samples at inference time; 
other methods are inductive.}
\label{tab:zsl_ntu}
\resizebox{0.9\linewidth}{!}{
\begin{tabular}{lccccc}
\toprule
\multirow{2}{*}{Method} & \multirow{2}{*}{Venue} & \multicolumn{2}{c}{NTU-60} & \multicolumn{2}{c}{NTU-120} \\
\cmidrule(lr){3-4} \cmidrule(lr){5-6}
 &  & 55/5 & 48/12 & 110/10 & 96/24 \\
\midrule
DeViSE~\cite{frome2013devise}      & NeurIPS'13  & 60.72 & 24.51 & 47.49 & 25.74 \\
RelationNet~\cite{jasani2019skeleton} & ICCV'19  & 40.12 & 30.06 & 52.59 & 29.06 \\
ReViSE~\cite{hubert2017learning}    & ICCV'17   & 53.91 & 17.49 & 55.04 & 32.89 \\
JPoSE~\cite{wray2019fine}       & ICCV'19   & 64.82 & 28.75 & 51.93 & 32.44 \\
CADA-VAE~\cite{schonfeld2019generalized} & CVPR'19 & 76.84 & 28.96 & 59.34 & 33.57 \\
SynSE~\cite{gupta2021syntactically}   & ICIP'21   & 75.81 & 33.30 & 62.69 & 38.70 \\
SMIE~\cite{zhou2023zero}        & MM'23    & 77.98 & 40.18 & 65.74 & 45.30 \\
SA-DVAE~\cite{li2024sa}       & ECCV'24   & 82.37 & 41.38 & 68.77 & 46.12 \\
DVTA~\cite{kuang2025zero}       & PR'25   & 79.28 & 44.14 & 74.89 & 51.81 \\
STAR~\cite{chen2024fine}        & MM'24    & 81.40 & 45.10 & 63.30 & 44.30 \\
GZSSAR~\cite{li2023multi}      & ICIG'23   & 83.30 & 49.80 & 72.00 & 60.70 \\
InfoCPL~\cite{xu2025information}    & TMM'25   & 85.91 & 53.32 & 74.81 & 60.05 \\
PURLS~\cite{zhu2024part}       & CVPR'24   & 79.23 & 40.99 & 71.95 & 52.01 \\
ScoPLe~\cite{zhu2025semantic}     & CVPR'25   & 84.10 & 52.96 & 74.53 & 52.17 \\
Neuron~\cite{chen2025neuron}      & CVPR'25   & 86.90 & 62.70 & 71.50 & 57.10 \\
FS-VAE~\cite{wu2025frequency}     & ICCV'25   & 86.90 & 57.20 & 74.40 & 62.50 \\
TDSM~\cite{do2025bridging}      & ICCV'25   & 86.49 & 56.03 & 74.15 & 65.06 \\
SC~\cite{zhu2026boosting}  & NeurIPS'25 & \textbf{89.41}  & 47.83  & 74.29 & 53.14\\
Flora~\cite{chen2025learning}     & CVPRF'26    & 86.30 & \underline{65.30} & \underline{79.60} & \underline{66.40} \\
\midrule
\textbf{GenPrior} & — & \underline{87.98} & \textbf{73.43}~{\scriptsize$\uparrow$8.1} & \textbf{90.44}~{\scriptsize$\uparrow$10.8} & \textbf{81.39}~{\scriptsize$\uparrow$14.9}  \\
\bottomrule
\end{tabular}
}
\end{table}

\begin{table}[t]
\centering
\caption{Evaluation under random split settings on the NTU-60, NTU-120, and PKU-MMD.}
\label{tab:zsl_additional}
\resizebox{0.85\linewidth}{!}{
\begin{tabular}{lcccc}
\toprule
\multirow{2}{*}{Method} & \multirow{2}{*}{Venue} & NTU-60 & NTU-120 & PKU-MMD \\
\cmidrule(lr){3-3} \cmidrule(lr){4-4} \cmidrule(lr){5-5}
 & & 55/5 & 110/10 & 46/5 \\
\midrule
ReViSE~\cite{hubert2017learning}  & ICCV'17  & 60.9  & 44.9  & 59.3 \\
JPoSE~\cite{wray2019fine}     & ICCV'19  & 59.4  & 46.7  & 57.2 \\
CADA-VAE~\cite{schonfeld2019generalized} & CVPR'19 & 61.8 & 45.2 & 60.7 \\
SynSE~\cite{gupta2021syntactically} & ICIP'21  & 64.2  & 47.3  & 60.8 \\
SMIE~\cite{zhou2023zero}      & MM'23   & 65.1  & 46.4  & 60.8 \\
SA-DVAE~\cite{li2024sa}      & ECCV'24  & 84.2  & 50.7  & 66.5 \\
ScoPLe~\cite{zhu2025semantic}   & CVPR'25  & 83.7  & 53.3  & 71.4 \\
TDSM~\cite{do2025bridging}    & ICCV'25  & 88.9  & 69.5  & 70.8 \\
SC~\cite{zhu2026boosting} & NeurIPS'25 & \underline{89.9}  & 56.2  & 71.1 \\
Flora~\cite{chen2025learning}   & CVPRF'26  & 88.6  & \underline{71.2}  & \underline{71.6} \\
\midrule
\textbf{GenPrior} & — & \textbf{96.3}~{\scriptsize$\uparrow$6.4} & \textbf{84.7}~{\scriptsize$\uparrow$13.5} & \textbf{74.7}~{\scriptsize$\uparrow$3.1} \\
\bottomrule
\end{tabular}
}
\end{table}

\begin{table*}[t]
\centering
\caption{Generalized zero-shot learning results (\%) on NTU-60 and NTU-120. \textit{S}, \textit{U}, and \textit{H} denote seen accuracy, unseen accuracy, and harmonic mean, respectively.}
\label{tab:gzsl}
\resizebox{0.7\linewidth}{!}{
\begin{tabular}{lc|ccc|ccc|ccc|ccc}
\toprule
\multirow{3}{*}{Method} & \multirow{3}{*}{Venue} & \multicolumn{6}{c|}{NTU-60} & \multicolumn{6}{c}{NTU-120} \\
\cmidrule(lr){3-8} \cmidrule(lr){9-14}
 & & \multicolumn{3}{c|}{55/5 Split} & \multicolumn{3}{c|}{48/12 Split} & \multicolumn{3}{c|}{110/10 Split} & \multicolumn{3}{c}{96/24 Split} \\
\cmidrule(lr){3-5} \cmidrule(lr){6-8} \cmidrule(lr){9-11} \cmidrule(lr){12-14}
 & & \textit{S} & \textit{U} & \textit{H} & \textit{S} & \textit{U} & \textit{H} & \textit{S} & \textit{U} & \textit{H} & \textit{S} & \textit{U} & \textit{H} \\
\midrule
ReViSE~\cite{hubert2017learning}  & ICCV'17  & 40.8 & 50.2 & 45.0 & 21.8 & 14.8 & 17.6 & 0.6 & 14.5 & 1.1 & 3.4 & 1.5 & 2.1 \\
JPoSE~\cite{wray2019fine}     & ICCV'19  & 66.5 & 53.5 & 59.3 & 28.6 & 18.7 & 22.6 & 53.6 & 11.6 & 19.1 & 41.0 & 3.8 & 6.9 \\
CADA-VAE~\cite{schonfeld2019generalized} & CVPR'19 & 56.1 & 56.0 & 56.0 & 50.4 & 25.0 & 33.4 & 50.2 & 43.9 & 46.8 & 48.3 & 27.5 & 35.1 \\
SynSE~\cite{gupta2021syntactically} & ICIP'21  & 51.3 & 47.4 & 49.2 & 44.1 & 22.9 & 30.1 & 57.3 & 43.2 & 49.5 & 48.1 & 32.9 & 39.1 \\
GZSSAR~\cite{li2023multi}    & ICIG'23  & 71.7 & 66.2 & 68.8 & 58.8 & 40.0 & 47.6 & 46.8 & \underline{68.3} & 55.6 & 56.8 & 48.6 & 52.4 \\
SA-DVAE~\cite{li2024sa}     & ECCV'24  & 62.8 & 70.8 & 66.3 & 50.2 & 36.9 & 42.6 & 61.1 & 59.8 & 60.4 & 58.8 & 35.8 & 44.5 \\
STAR~\cite{chen2024fine}      & MM'24   & 69.0 & 69.9 & 69.4 & 62.7 & 37.0 & 46.6 & 59.9 & 52.7 & 56.1 & 51.2 & 36.9 & 42.9 \\
ScoPLe~\cite{zhu2025semantic}   & CVPR'25  & 69.6 & 71.9 & 70.8 & 54.5 & \textbf{61.8} & 57.9 & 63.5 & 61.1 & 62.3 & 53.3 & 51.2 & 52.2 \\
Neuron~\cite{chen2025neuron}    & CVPR'25  & 69.1 & 73.8 & 71.4 & 61.6 & 56.8 & 59.1 & \textbf{67.6} & 59.5 & 63.3 & \textbf{67.5} & 44.4 & 53.6 \\
FS-VAE~\cite{wu2025frequency}   & ICCV'25  & \textbf{77.0} & 74.5 & \underline{75.7} & 56.2 & 48.6 & 52.1 & 59.2 & 67.9 & 63.3 & 57.8 & \underline{51.9} & \underline{54.7} \\
SC~\cite{zhu2026boosting} & NeurIPS'25 & 64.6  & \underline{79.2}  & 71.2  & 53.2  & 44.4  & 48.4  & 62.2  & 65.8  & 64.0  & \underline{59.9}  & 44.1  & 50.8 \\
Flora~\cite{chen2025learning}   & CVPRF'26  & \underline{75.9} & 78.8 & \textbf{77.4} & \underline{63.7} & 57.5 & \underline{60.5} & 66.2 & 66.0 & \underline{66.1} & 55.9 & 50.7 & 53.2 \\
\midrule
\textbf{GenPrior} & — & 73.5 & \textbf{80.0} & \underline{76.6} & \textbf{65.5} & \underline{61.4} & \textbf{63.4} & \underline{67.1} & \textbf{75.2} & \textbf{70.9} & 56.5 & \textbf{55.4} & \textbf{55.9} \\
\bottomrule
\end{tabular}
}
\end{table*}

\subsection{Comparison with State-of-the-Art}
\textbf{ZSL Evaluation on Standard Splits:}
Table~\ref{tab:zsl_ntu} reports the zero-shot recognition results on NTU-60 and NTU-120. Our proposed GenPrior achieves state-of-the-art performance. We attain accuracies of 90.44\%  (+10.8\%) and 81.39\%  (+14.9\%) on the NTU-120 110/10 and 96/24 splits, outperforming the leading inductive method Flora~\cite{chen2025learning}. Furthermore, our Generative Prototype Refinement module operates in a test-time adaptation setting. Compared to SC~\cite{zhu2026boosting}, a recent baseline also utilizing TTA, GenPrior achieves a substantial +28.2\% gain (81.39\% vs. 53.14\%) on the NTU-120 96/24 split. Crucially, it is noteworthy that the performance margin expands drastically as the proportion of unseen classes increases. This trend empirically demonstrates that when the semantic space becomes highly ambiguous due to a large number of unseen categories, pure textual prototypes suffer from a severe semantic-kinematic gap. In contrast, our generation enhanced semantic features serve as essential physically grounded anchors, providing decisive discriminative cues that pure text inherently lacks.

\noindent \textbf{ZSL Evaluation on Random Splits:}
Table~\ref{tab:zsl_additional} presents the evaluation results under the random split setting on the NTU-60, NTU-120, and PKU-MMD datasets, verifying the robustness of our method across diverse category combinations. Our approach outperforms all counterparts across the three datasets, achieving absolute improvements of 6.4\%, 13.5\%, and 3.1\%, respectively. This consistent enhancement indicates that the effectiveness of incorporating generative motion priors is not biased toward specific category splits and generalizes well to different datasets.

\noindent \textbf{GZSL Evaluation:}
Table~\ref{tab:gzsl} reports the results of Generalized Zero-Shot Learning, where the search space encompasses both seen and unseen classes. On the 55/5 split of NTU-60, while maintaining a competitive Harmonic mean (H), our method significantly boosts the Unseen accuracy (U) to 80.0\%. On the 48/12 split, we achieve state-of-the-art performance in both Seen accuracy (65.5\%) and Harmonic mean (63.4\%). These results demonstrate that our Generative Prototype Refinement module effectively alleviates the seen-class bias inherent in GZSL, as the generation-enhanced prototypes endow the unseen classes with a much more competitive representation in the feature space.

\begin{table}[htbp]
  \centering
  \small
  \caption{Ablation study of components on NTU-60 and NTU-120. GESF denotes Generation Enhanced Semantic Features, and GPR denotes Generative Prototype Refinement.}
  \label{tab:ablation}
  \begin{tabular}{cccccc}
    \toprule
    \multirow{2}{*}{GESF} & \multirow{2}{*}{GPR} & \multicolumn{2}{c}{NTU-60} & \multicolumn{2}{c}{NTU-120} \\
    \cmidrule(lr){3-4} \cmidrule(lr){5-6}
    & & 55/5 & 48/12 & 110/10 & 96/24 \\
    \midrule
    \ding{55} & \ding{55} & 84.96 & 57.89 & 77.59 & 65.16 \\
    \ding{55} & \ding{51} & 85.47 & 64.65 & 87.47 & 77.35 \\
    \ding{51} & \ding{55} & 86.73 & 61.44 & 78.80 & 69.17 \\
    \ding{51} & \ding{51} & \textbf{87.98} & \textbf{73.43} & \textbf{90.44} & \textbf{81.39} \\
    \bottomrule
  \end{tabular}
\end{table}

\begin{table}[htbp]
\centering
\small
\caption{Ablation study of GESF on NTU-60 and NTU-120.}
\label{tab:GESF_ablation}
\setlength{\tabcolsep}{4pt}
\begin{tabular}{ccccccc}
\toprule
\multirow{2}{*}{Proto $\hat{\mu}_y$} & \multirow{2}{*}{Disp $\hat{\sigma}_y$} & \multirow{2}{*}{Text $e_y$} & \multicolumn{2}{c}{NTU-60} & \multicolumn{2}{c}{NTU-120} \\
\cmidrule(lr){4-5} \cmidrule(lr){6-7}
 & & & 55/5 & 48/12 & 110/10 & 96/24 \\
\midrule
\ding{55} & \ding{55} & \ding{55} & 85.47 & 64.65 & 87.47 & 77.35 \\
\ding{51} & \ding{55} & \ding{55} & 86.15 & 63.52 & 88.31 & 76.91 \\
\ding{51} & \ding{51} & \ding{55} & 87.24 & 70.18 & 89.67 & 80.25 \\
\ding{51} & \ding{55} & \ding{51} & 86.83 & 68.27 & 89.12 & 79.58 \\
\ding{51} & \ding{51} & \ding{51} & \textbf{87.98} & \textbf{73.43} & \textbf{90.44} & \textbf{81.39} \\
\bottomrule
\end{tabular}
\end{table}

\begin{table}[htbp]
\centering
\small
\caption{Effect of different T2M backbones on ZSL accuracy.}
\label{tab:t2m_model}
\setlength{\tabcolsep}{5pt}
\begin{tabular}{lcccc}
\toprule
\multirow{2}{*}{T2M Model} & \multicolumn{2}{c}{NTU-60} & \multicolumn{2}{c}{NTU-120} \\
\cmidrule(lr){2-3} \cmidrule(lr){4-5}
 & 55/5 & 48/12 & 110/10 & 96/24 \\
\midrule
MLD~\cite{chen2023executing}            & 87.42 & 72.15 & 89.78 & 80.63 \\
ReMoDiffuse~\cite{zhang2023remodiffuse}    & 87.65 & 72.87 & 90.11 & 80.95 \\
StableMoFusion~\cite{huang2024stablemofusion} & \textbf{87.98} & \textbf{73.43} & \textbf{90.44} & \textbf{81.39} \\
\bottomrule
\end{tabular}
\end{table}

\begin{figure}[t]
    \centering
    \includegraphics[width=\columnwidth]{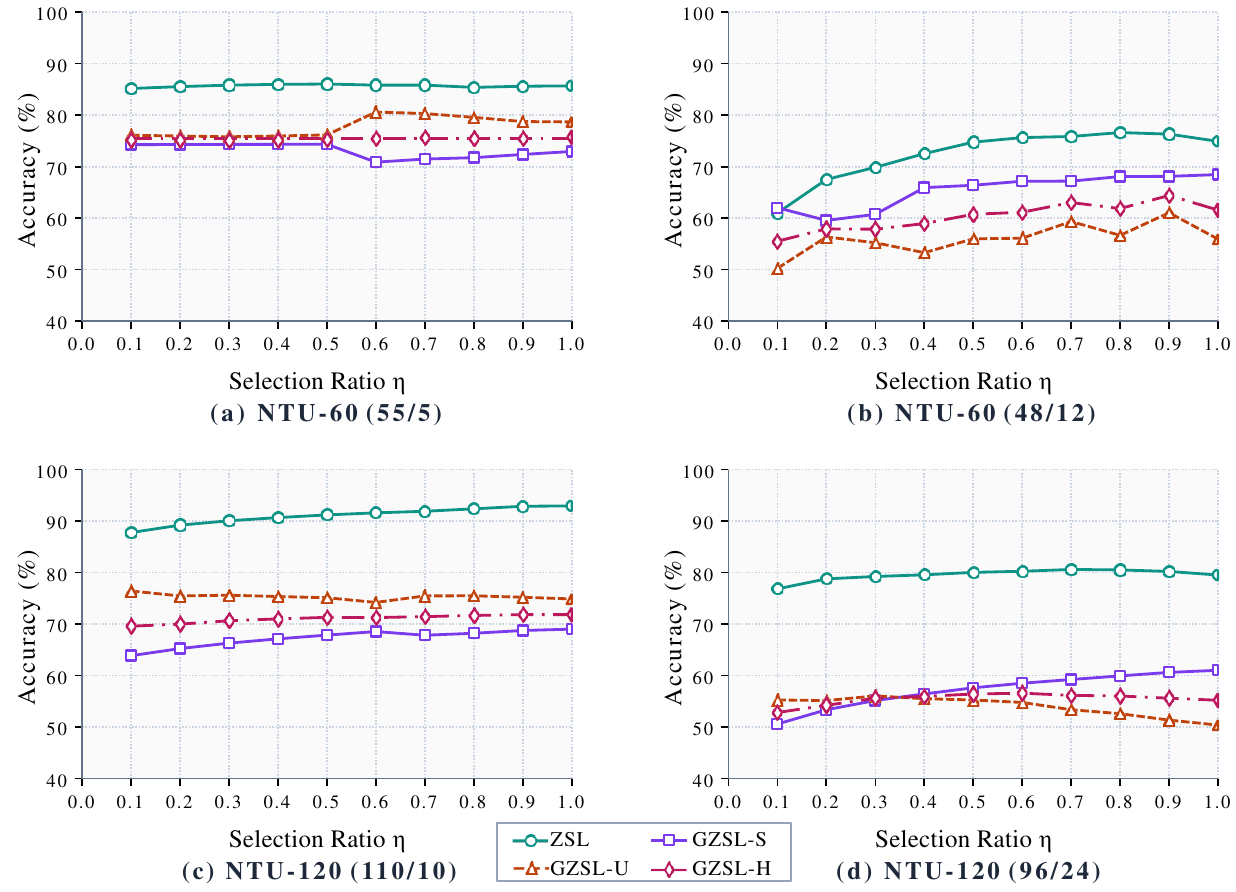}
    \caption{Sensitivity analysis of selection ratio $\eta$ in GPR across four evaluation splits. ZSL accuracy and GZSL metrics (S, U, H) are reported. The model exhibits stable performance across a wide range of $\eta$, with optimal results typically achieved around $\eta = 0.5$.}
    \label{fig:eta_sensitivity}
\end{figure}

\subsection{Ablation Studies}
\noindent\textbf{Effect of Core Components:}
To verify the contribution of each key component, we conduct ablation experiments on four splits of NTU-60 and NTU-120, as summarized in Table~\ref{tab:ablation}. The baseline in the first row relies solely on textual features for cross-modal VAE alignment without any generative prior. Introducing GPR alone yields consistent improvements across all splits, with a notable gain of 12.19\% on the 96/24 split. This demonstrates the effectiveness of adaptive prototype refinement. Introducing GESF alone also improves ZSL accuracy by injecting kinematic cues into the semantic space. Most importantly, combining GESF and GPR produces a strong synergistic effect. On the challenging 48/12 and 96/24 splits, where the unseen class set is large and semantically ambiguous, the full model achieves gains of 15.54\% and 16.23\% over the baseline, respectively. This confirms that the kinematically grounded representations from GESF provide reliable anchors and substantially improve the quality of GPR's confidence-based sample mining.

\noindent\textbf{Effect of Generation Enhanced Semantic Features (GESF):}
Table~\ref{tab:GESF_ablation} further examines the necessity of each input signal in the GESF gating mechanism. All variants include GPR to isolate the effect of GESF design choices. When the generative prototype $\hat{\mu}_y$ is directly injected without any gating, corresponding to the second row with $g_y{=}1$, performance on the harder splits degrades. For instance, the accuracy drops from 64.65\% to 63.52\% on the 48/12 split. This confirms that ungated injection introduces domain artifacts that harm recognition. Incorporating the dispersion vector $\hat{\sigma}_y$ as a gating signal, as shown in the third row, recovers and substantially surpasses the no-prior baseline. This indicates that generation uncertainty is an effective indicator for controlling injection strength. Adding the textual feature $e_y$ as a joint input, corresponding to the fourth row, provides complementary gains by enabling the gate to assess semantic-kinematic compatibility. The full three-input gate in the last row achieves the best results across all splits, validating that jointly conditioning on generation quality and cross-modal complementarity yields the most effective semantic fusion.

\noindent\textbf{Effect of Selection Ratio $\eta$ in Generative Prototype Refinement (GPR):}
Figure~\ref{fig:eta_sensitivity} analyzes the sensitivity of GPR to the high-confidence selection ratio $\eta$ across four evaluation splits. Both ZSL accuracy and GZSL metrics (S, U, H) remain relatively stable over a wide range of $\eta$ from 0.1 to 1.0, indicating that GPR is robust to this hyperparameter in practice. When $\eta$ is very small, the calibrated prototypes lack sufficient intra-class coverage due to the limited number of selected samples. When $\eta$ approaches 1.0, low-confidence samples near decision boundaries are inevitably included, leading to slight performance degradation. The optimal trade-off between support set purity and distribution coverage is generally achieved at $\eta \in [0.5, 0.7]$.

\begin{figure*}[htbp]
    \centering
    \includegraphics[width=0.95\textwidth]{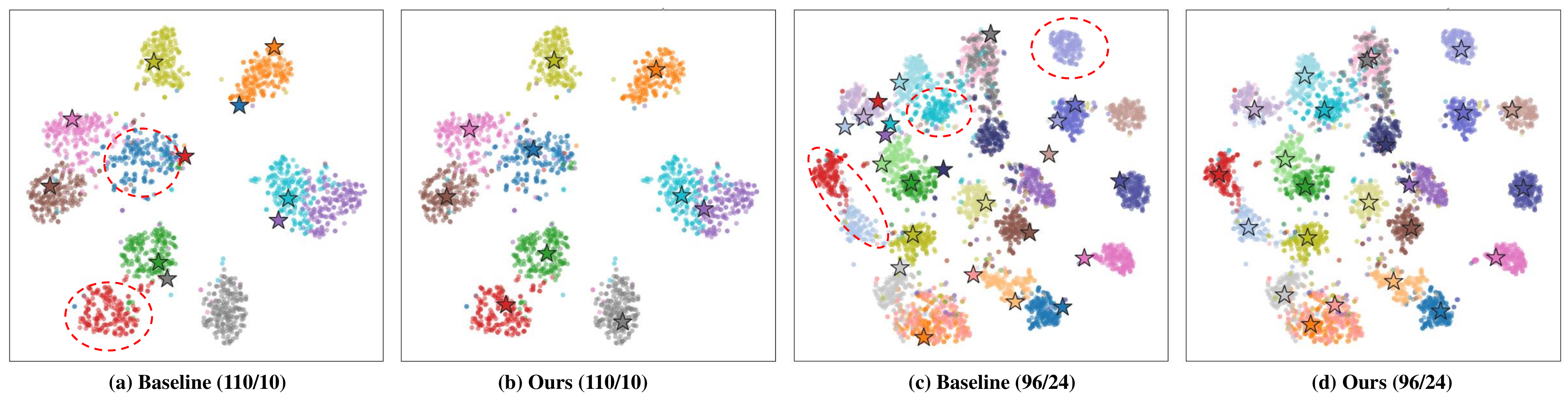}
    \caption{t-SNE visualization of unseen-class test samples (dots) and semantic prototypes (stars) on NTU-120. Red dashed circles highlight cases where baseline prototypes deviate significantly from their corresponding sample clusters.}
    \label{fig:t-sne}
\end{figure*}

\begin{figure}[htbp]
    \centering
    \includegraphics[width=0.8\columnwidth]{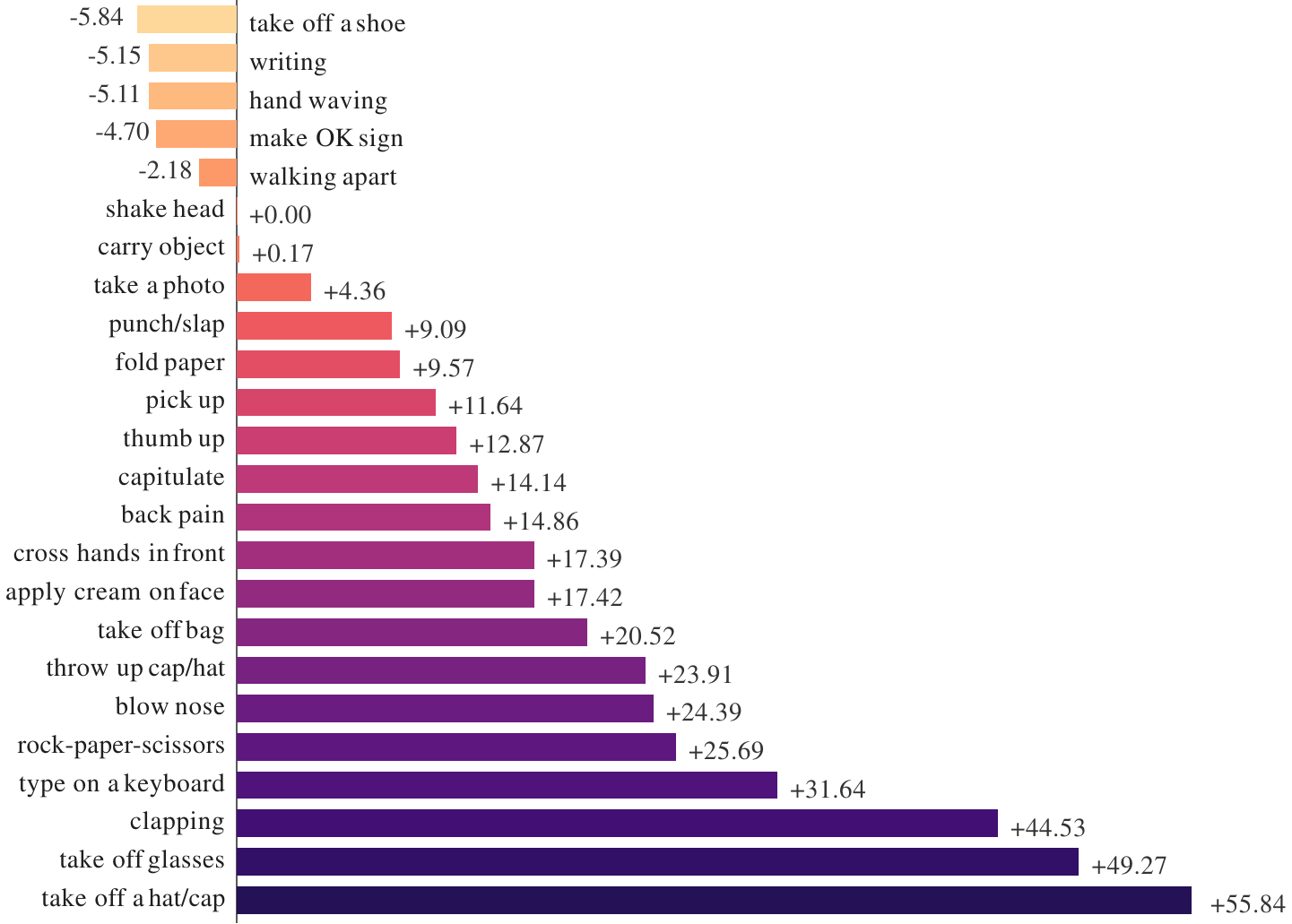}
    \caption{Per-class accuracy change (\%) after applying GPR on NTU-120 (96/24). Positive values indicate improvement.}
    \label{fig:per-class}
\end{figure}

\noindent\textbf{Effect of T2M Models:}
Table~\ref{tab:t2m_model} investigates the impact of different Text-to-Motion (T2M) generative models, used as prior sources, on the zero-shot recognition performance. We compare three representative models, namely MLD, ReMoDiffuse, and StableMoFusion. The results show that StableMoFusion achieves the best performance across all evaluated splits. This advantage can be mainly attributed to its stronger cross-modal alignment capability, which enables the generation of motions that better conform to the underlying text semantics. Nevertheless, although these three generative models differ in their underlying architectures and denoising mechanisms, they produce highly consistent recognition performance across all splits, with performance variations remaining within 1.3\%. This observation indicates that GenPrior is not tied to any specific generative model, but can benefit from any T2M source capable of providing kinematically plausible motion priors.

\subsection{Qualitative Results}

\noindent\textbf{Feature Space Visualization:}
To rigorously demonstrate the improvement in unseen-class prototype quality achieved by GenPrior, we present t-SNE visualizations on the NTU-120 dataset with 110/10 and 96/24 splits, as shown in Figure~\ref{fig:t-sne}. In the visualization, dots represent test samples and stars denote class prototypes. In the baseline model, shown in subfigures (a) and (c), prototypes derived solely from textual encodings are substantially misaligned with their corresponding sample clusters. The red dashed circles highlight representative cases where prototypes deviate from cluster centers or overlap with neighboring class regions. This pattern reveals the semantic-kinematic gap in the feature space. By contrast, the GenPrior model, shown in subfigures (b) and (d), produces prototypes that align closely with the centers of their respective clusters, and inter-class boundaries are more clearly separated. This improvement is particularly evident in the 96/24 split, where the number of unseen classes is larger. The baseline exhibits severe prototype misalignment and increased inter-class confusion, whereas GenPrior consistently achieves reliable prototype-cluster alignment.

\noindent\textbf{Class-wise Performance:}
To further analyze the class-level impact of GPR, Figure~\ref{fig:per-class} illustrates the per-class accuracy changes before and after GPR calibration on the NTU-120 96/24 split. GPR leads to substantial improvements for most unseen classes. The largest gains are observed in categories such as ``take off a hat or cap'' (+55.84\%) and ``take off glasses'' (+49.27\%). These actions involve fine-grained object interactions, where subtle kinematic differences are difficult to capture using textual prototypes alone. GPR addresses this limitation by leveraging high-confidence real samples to refine decision boundaries. A small number of classes, including ``take off a shoe'' (-5.84\%) and ``writing'' (-5.15\%), show slight performance degradation. Overall, GPR improves performance on 19 out of 24 unseen classes, demonstrating the effectiveness and general applicability of the generative-anchor-guided confidence mining strategy.

\section{Conclusion}

We present GenPrior, a novel framework for zero-shot skeleton-based action recognition that incorporates pre-trained Text-to-Motion models as external motion prior sources. GenPrior addresses the semantic-kinematic gap by enriching semantic representations with kinematic cues derived from motion priors and refining unseen-class prototypes using high-confidence samples. Extensive experiments on NTU-60, NTU-120, and PKU-MMD demonstrate consistent improvements over existing approaches under both ZSL and GZSL settings. We believe that exploiting generative motion priors for 
discriminative skeleton understanding opens a promising direction that can be extended to few-shot action recognition and cross-dataset transfer scenarios.

\begin{acks}
This work was supported in part by the National Natural Science Foundation of China (62302093, 52441503), the Natural Science Foundation of Jiangsu Province (BK20230833), and the CIPS-SMP-Zhipu Large Model Fund. We thank the Big Data Computing Center of Southeast University for providing the facility support on the numerical calculations.
\end{acks}

\bibliographystyle{ACM-Reference-Format}
\bibliography{sample-base}

\end{document}